# Prompt engineering does not universally improve Large Language Model performance across clinical decision-making tasks

Mengdi Chai [1,2] and Ali R. Zomorrodi [2,3,4*]

[1] Department of Biostatistics, Harvard School of Public Health, Boston, MA

[2] Mucosal Immunology and Biology Research Center, Department of Pediatrics, Massachusetts General Hospital, Boston, MA

[3] Harvard Medical School, Boston, MA

[4] Broad Institute of MIT and Harvard, Boston, MA

[*]Corresponding author

E-mail: azommorrodi@mgh.harvard.edu

## Abstract

In this study, we evaluated the effects of prompt engineering on the clinical reasoning performance of ChatGPT-4o, Gemini 2.5 Flash, and Llama 3.3 70B across five core clinical decision-making tasks using 36 clinical cases.  Prompt engineering using two variants of MedPrompt improved performance on relevant diagnostic testing, the most challenging task at baseline, but produced mixed and sometimes adverse effects elsewhere, indicating that prompt engineering is not a one-size-fits-all solution.

## Introduction

Clinical decision-making is recognized as a cognitively demanding process in safe and effective patient care. Yet diagnostic and therapeutic errors remain alarmingly prevalent [1]. Large Language Models (LLMs) have emerged as a potentially transformative paradigm for clinical decision support. Recent studies have highlighted this potential by documenting their strong performance on standardized medical knowledge benchmarks such as USMLE and MedQA, achieving expert-level accuracy  or even surpassing the performance of trained physicians [2,3]. Prompt engineering strategies—such as chain-of-thought (CoT) reasoning and few-shot prompting—have been shown to enhance LLM reasoning in both medical and non-medical tasks [4,5], with the MedPrompt framework as a prominent example [6].

However, success on knowledge-based examinations does not translate to effective performance in real-world clinical reasoning, which requires the integration of patient-specific data, prioritization under uncertainty, and the timing and arrangement of diagnostic test and therapeutic interventions [7-10]. Moreover, approaches such as MedPrompt have been tested almost exclusively on OpenAI models like ChatGPT [6], and while these methods are model agnostic, their generalizability to other LLMs has not been explored. This underscores the pressing need for systematic evaluations that assess how different prompting strategies influence LLM performance across the full spectrum of clinical workflow. In this study, we aimed to address this gap by systematically evaluating knowledge recall and the clinical reasoning performance of three state-of-the-art LLMs—ChatGPT-4o (OpenAI), and Gemini 2.5 Flash (Google), and Llama 3.3 70B (Meta)—across five key stages of sequential clinical decision-making using both baseline prompting and prompt engineering approaches.

## Results

A total of 36 publicly available clinical cases were extracted from the Merck Sharpe & Dohme (MSD) Clinical Manual [11]. These clinical vignettes are designed to reflect real-world clinical scenarios that healthcare professionals commonly encounter and involve a structured clinical reasoning workflow that begins with patient's history of present and past illness (HPI), review of systems (ROS), and physical examination (PE); it then proceeds through five sequential decision-making tasks: differential diagnosis, essential immediate steps, relevant diagnostic testing, final diagnosis, and treatment recommendations (**Figure 1**). For each decision-making task, LLMs were provided with progressively richer clinical input: initially, HPI, ROS, and PE for differential diagnosis; and for the remaining tasks, HPI, ROS, and PE along with the ground truth for all prior tasks in the sequence. This ensured the model had access to all pertinent context as it moved through the clinical reasoning sequence.

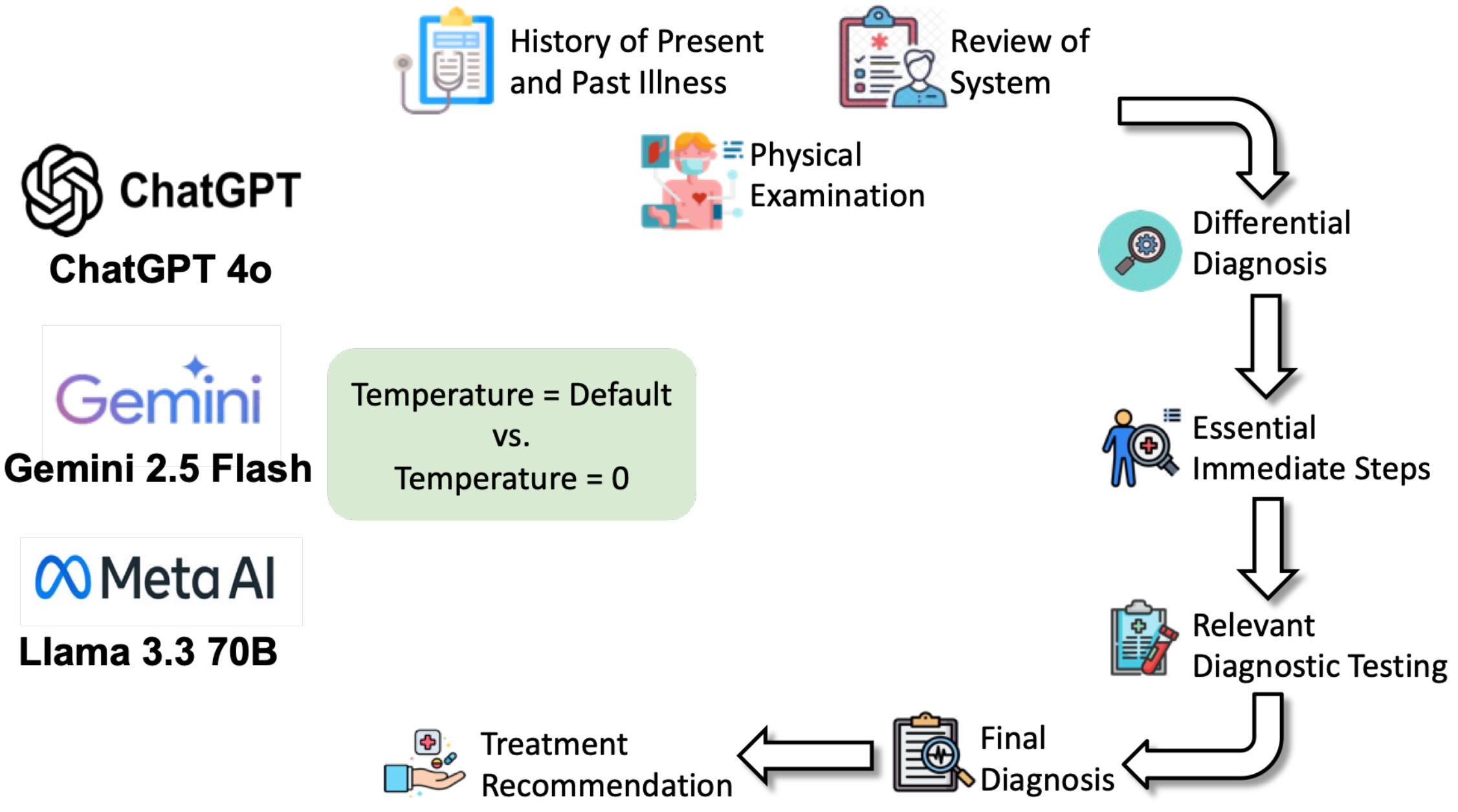

**Figure 1**. Workflow for assessing LLM performance across diverse clinical decision-making tasks. For each task, the HPI, ROS, and PE results, the ground truth answers from all previous tasks, and the current set of answer choices relevant to the task were provided as inputs to the LLM.

We first comprehensively evaluated each LLM's out-of-the-box performance profile across the five clinical reasoning tasks using basic prompting in order to better understand their inherent strengths and weaknesses in a more realistic clinical context. To this end, relevant patient information for each reasoning task was presented to the models without additional prompt engineering strategies. We further investigated whether the performance may differ between the default temperature for each model and zero temperature settings. The findings of this analysis are summarized in **Figure 2**. LLMs showed marked task dependence across the clinical reasoning workflow. All three models achieved near-perfect performance for final diagnosis, the lowest performance for relevant diagnostic testing, and moderate to moderately high accuracy for differential diagnosis, essential immediate steps, and treatment recommendation. Across models, Gemini 2.5 Flash consistently matched or outperformed its peers in all tasks. Although modest model-specific differences were observed between default and zero temperature settings, none were statistically significant (Wilcoxon Signed Rank test, $p < 0.05$); directionally, Gemini 2.5 Flash tended to perform slightly better at zero temperature, Llama 3.3 70B was largely insensitive to temperature setting, and ChatGPT-4o showed mixed temperature-dependent patterns across tasks.

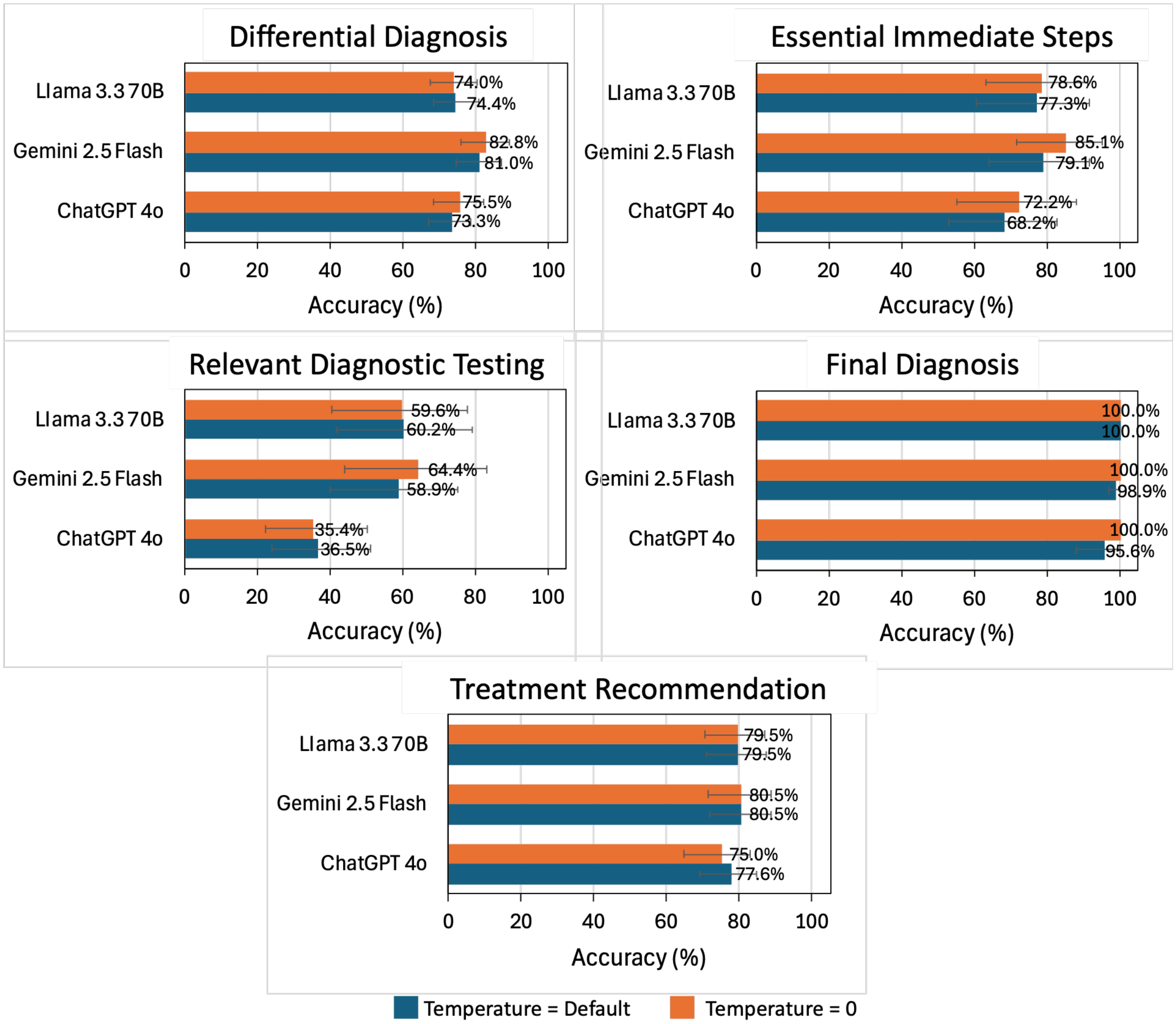

**Figure 2**. Evaluation of the three LLMs' out-of-the-box clinical reasoning performance under default and zero temperature settings using baseline prompting. For each vignette and task, performance was evaluated in three independent runs and averaged to obtain a vignette-level score; task-level performance reflects the mean across vignettes. Error bars denote 95% confidence intervals (see **Methods**). Although slight differences were observed in LLM performance between the default and zero temperature settings across all five clinical reasoning tasks and for all three models, statistical analysis confirmed that none of the observed differences in model accuracy between temperature settings were statistically significant (Wilcoxon Signed Rank test, $p < 0.05$; **Supplementary File 1**). To partially separate omission-driven versus over-selection errors, we additionally reported precision and F1 score alongside the overlap-based accuracy (which corresponds to recall) in **Supplementary File 1**.

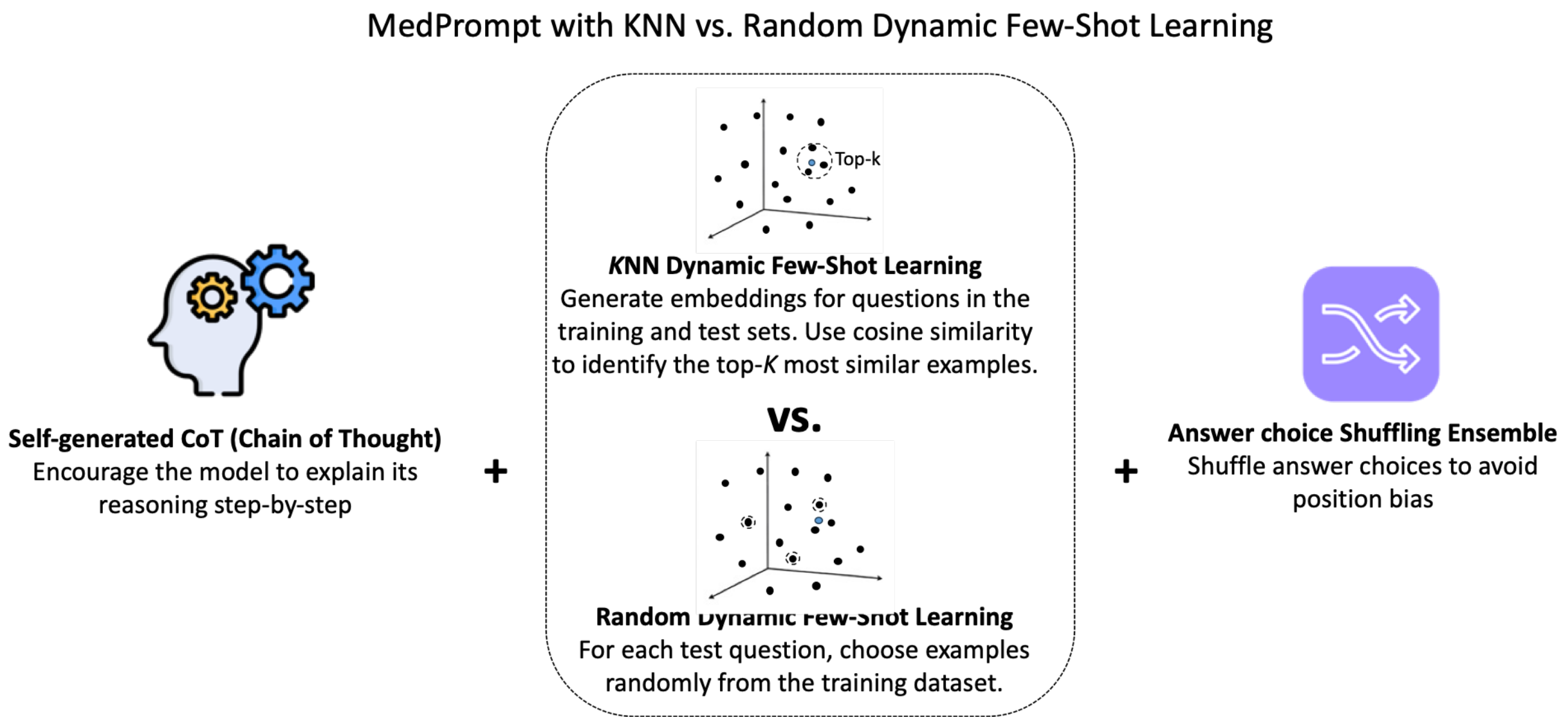

**Figure 3**. Overview of structured prompt engineering strategies evaluated in this study across clinical reasoning tasks. The MedPrompt framework [6] was assessed using both the original $K$NN and random dynamic few-shot selection methods.

Following our baseline evaluation, we sought to determine whether prompt engineering could further enhance the clinical reasoning capabilities of the LLMs. To this end, we leveraged prompt engineering strategies from MedPrompt 6, a structured prompt engineering framework that integrates self-generated CoT reasoning with dynamic few-shot learning and answer choice shuffling ensemble. In this study, we used the same self-generated CoT and answer choice shuffling approaches as in MedPrompt 6. For dynamic few-shot learning, however, we explored two different approaches. The first utilized a targeted dynamic few-shot selection scheme based on the K nearest neighbors (KNN), where K examples are chosen based on their semantic similarity to the test question, as originally proposed in MedPrompt 6. The second approach substituted this with random dynamic few-shot selection, where K exemplars were chosen at random from the training set (Figure 3). We excluded final diagnosis from our analysis as all three LLMs showed near-complete accuracy for this task using basic prompting (Figure 2), which left little room for further improvement. The results of this analysis is shown in Figure 4. Neither MedPrompt variant produced uniform performance gains across models or tasks. Prompt engineering substantially improved performance for relevant diagnostic testing, the most challenging task in out-of-the-box evaluation across all three models, except for Gemini 2.5 Flash with random few-shot prompting. For example, the accuracy of ChatGPT 4o for relevant diagnostic testing surged from 30.8% to 50.7% with MedPrompt using KNN few-shots (a 19.9% improvement), whereas Llama 3.3 70B and Gemini 2.5 Flash showed 17.9% and 13.5% improvement, respectively.

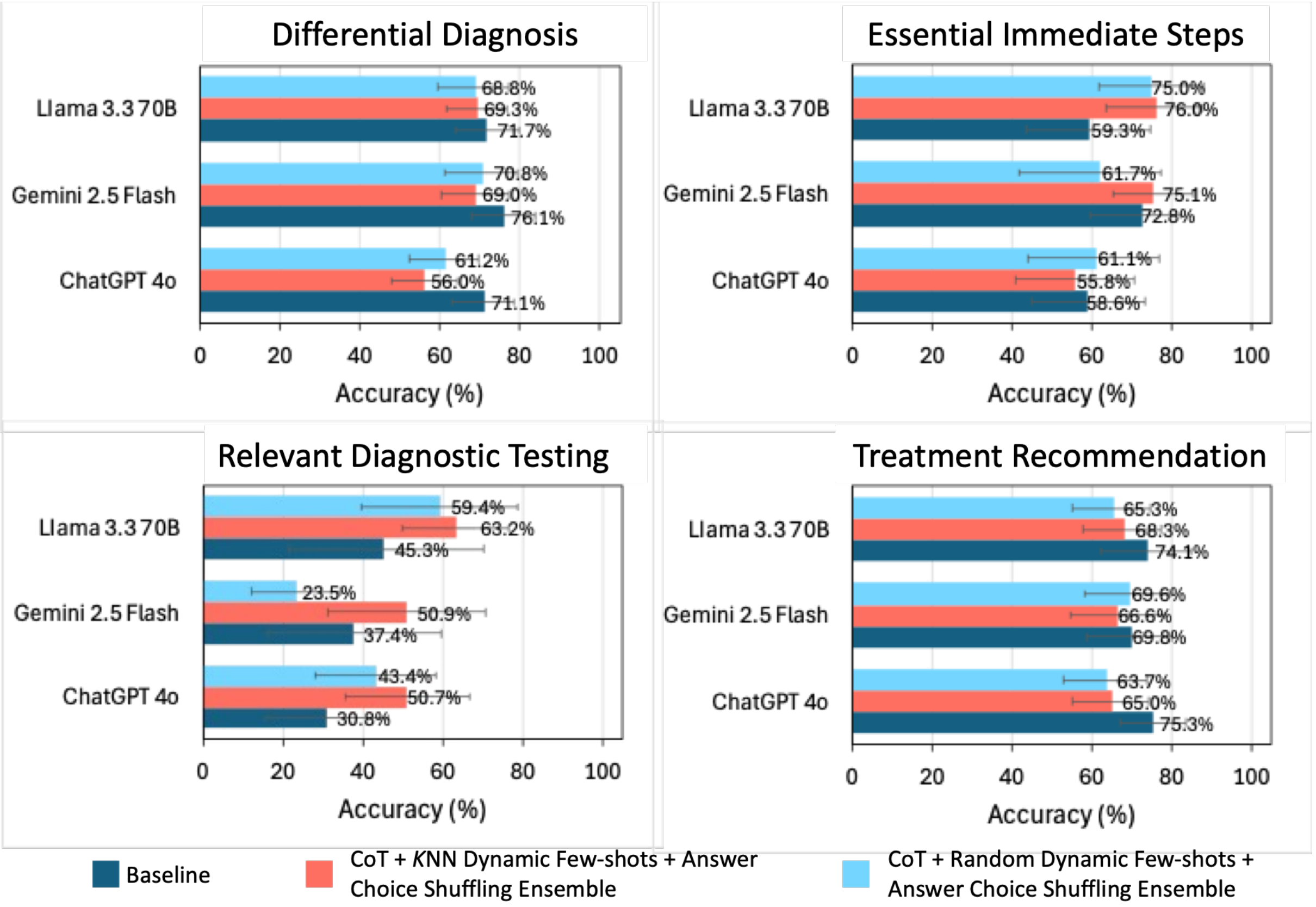

**Figure 4**. Effects of prompt engineering strategies on LLM performance across clinical decision-making tasks. Given that our out-of-the-bx performance analysis revealed no statistically significant performance differences between temperature settings, all prompt engineering evaluations were conducted using each model's default temperature. Prompt engineering was based on the MedPrompt framework [6], which integrates CoT reasoning, $K$NN dynamic few-shot learning, and answer choice shuffling assemble. In this study, we also explored a variant of MedPrompt in which $K$NN dynamic few-shot learning was substituted with random dynamic few-shot learning. For each task, approximately 30% of the cases were allocated to the training set (set of few-shot examples with self-generated CoT reasoning), while the remaining cases were reserved for model evaluation (see **Methods** for details). The LLM performance before and after applying prompt engineering was compared for the test dataset; the baseline for this analysis was the performance on the test dataset without prompt engineering (rather than performance on the entire dataset as in **Figure 2**). For each vignette, task, and prompt engineering strategy, performance was evaluated in three independent runs and averaged to obtain a vignette-level score. Task-level performance shows the mean across vignettes. Error bars indicate 95% confidence intervals (see **Methods**). Complementary performance metrics including precision and F1-scores are provided in **Supplementary File** 1. Only a small number of pairwise comparisons were statistically significant (Wilcoxon Signed Rank test, $p < 0.05$); full statistical analysis results are provided in **Supplementary File 1**.

However, prompt engineering using MedPrompt had mixed or even counterproductive effects elsewhere, including consistent performance declines in differential diagnosis and treatment recommendation for all or most models. For instance, MedPrompt with KNN dynamic few-shot prompting led to a sharp decline in ChatGPT 4o's differential diagnosis performance from 71.1% to 56.0%. Furthermore, KNN dynamic few-shot prompting did not consistently outperform random few-shot prompting, further indicating that the effect of prompt engineering was strongly task- and model-dependent.

## Discussion

Our findings indicate that LLM performance is shaped by the clinical task, the model, and the prompting strategy applied. The highly inconsistent and sometimes counterproductive effects of prompt engineering revealed that added reasoning structure by MedPrompt is beneficial only in selected settings. This motivates closer consideration of how task complexity, model behavior, prompting strategy, and deployment context shape LLM clinical reasoning performance.

**Task-dependent out-of-the-box LLM performance suggests differing reasoning demands across tasks:** Our analysis of out-of-the-box LLM performance revealed significant variability across clinical tasks (**Figure 2**). This performance spectrum could reflect the inherent cognitive demand and complexity of each task. The near-perfect accuracy in final diagnosis, for instance, can be likely attributed to the availability of a complete set of information to the model, including all prior ground truth and definitive diagnostic test results, which significantly constrains the diagnosis space and minimizes uncertainty. Conversely, relevant diagnostic testing likely proved the most difficult because it requires reasoning under uncertainty, weigh the potential diagnostic yield of multiple test options, and strategize the most efficient path to a diagnosis. This suggests that different stages of the clinical workflow place distinct reasoning demands on LLMs.

**Model accuracy was robust to temperature changes, though the choice remains critical for clinical applications:** Our initial rationale for including the zero temperature setting in our analysis was that reduced variability in model outputs might help ensure more consistent and reproducible clinical reasoning, which is particularly valuable in high-stakes healthcare environments. On the other hand, zero temperature may also compromise the performance as it could potentially favor more conservative or prototypical responses, impacting the breadth or creativity of the model's clinical reasoning. Examining both settings allowed us to explore whether certain clinical reasoning tasks benefit from more predictable, consistent responses (high determinism) versus allowing for a more flexible and broader range of potential solutions (greater breadth). Our analysis identified subtle model-specific trends between temperature settings, although

none of the observed differences were statistically significant. This suggests that model accuracy is mostly robust to this hyperparameter within the scope of the MSD dataset. However, the choice of temperature setting remains clinically significant. While the observed differences in accuracy were marginal in this benchmark, the degree of output determinism vs. variability has implications beyond performance alone. In particular, more stable responses can facilitate the auditability of model outputs across identical inputs, medico-legal accountability, and integration into clinical workflows. The decision to prioritize deterministic, reproducible outputs over more variable responses depends on the specific clinical task and intended use case and is a critical consideration for the safe and reliable deployment of these models in clinical workflows. This warrants further exploration across larger and more diverse datasets.

**Structured prompting can degrade performance by over-constraining a model's reasoning process:** A key finding of our study is that while MedPrompt's structured prompt engineering aims to guide models toward more systematic reasoning, it does not universally improve model performance, and in several cases, it can paradoxically be counterproductive, particularly when the prompts impose elaborate reasoning scaffolds or demand tight alignment between example semantics. Performance drops when subjected to prompt engineering were most evident for ChatGPT-4o; however, this effect is not unique to one model or one task; Llama 3.3 70B and Gemini 2.5 Flash also experienced marked performance degradation in treatment recommendation and differential diagnosis, respectively, with both prompting strategies (**Figure 4**). This indicates that even models that benefit from advanced prompting in some scenarios, can be impaired by it in others.

These observations suggest that, in some settings, imposing strict step-by-step reasoning and multi-example few-shot prompting may be counterproductive. Although the mechanism underlying this performance degradation under structured prompting cannot be resolved directly from our benchmark, a plausible explanation is that prompt length and context-window competition may lead to distraction from core task-relevant signals. Specifically, longer and more complex prompts—with added instructions, multiple examples, and mandated reasoning structure—consume limited context and can shift model attention toward generic or example-consistent patterns, thereby reducing its ability to focus on the clinically relevant details of the current vignette when selecting an answer. Emerging findings in the broader LLM similarly caution that forcing models into rigid, structured reasoning paths can undermine their natural reasoning process [12-14]. For example, both Kim et. al [12] and Shao et al [14] argue that CoT functions as a structural constraint that drives models to mimic reasoning patterns rather than engage in genuine, flexible, and novel inference pathways.

**Prompt engineering was most effective for tasks with greatest baseline difficulty:** Prompt engineering provided substantial benefits for the single most challenging task in out-of-the-box emulation: relevant diagnostic testing (**Figure 4**). Across all three models, both or at least one MedPrompt variant appeared to provide an effective structure aid for guiding the models' reasoning in a setting where their baseline capabilities were weakest. This suggests that when a model lacks a robust internal strategy for a high-uncertainty task, the explicit structure of a prompt could be highly beneficial.

**Semantic similarity alone may not optimize few-shot prompting:** Another notable observation was that MedPrompt with the more targeted $K$NN dynamic few-shots did not always outperform that with random dynamic few-shots (**Figure 4**). This finding indicates that relying solely on semantic similarity to select few-shot examples may not always yield the best results for model performance. The reason for this unexpected observation could be that the presumed advantage of closely matched examples can be offset in some cases by the loss of broader contextual diversity or alternative reasoning strategies. These mixed outcomes are in contrast to the consistent improvements reported in the original MedPrompt paper for medical knowledge exams [6] and highlight the intricate interplay between model architecture, the specific clinical reasoning task, and the chosen prompting strategy in real-world clinical scenarios.

**The study's findings should be interpreted in the context of several key limitations:** It is important to consider several limitations when interpreting our findings. First, this work is an exploratory benchmark conducted on a modest dataset of 36 clinical vignettes with uneven task-specific sample sizes, including a small sample size for essential immediate steps (n=10). The modest and uneven sample sizes across tasks reduce statistical power for task-specific comparisons and limit our ability to distinguish stable performance differences from sampling variability. These considerations reinforce the need for larger-scale evaluations. Second, the clinical cases were sourced from the MSD Manual, an evidence-based educational resource with curated, well-structured narratives and answer sets. Although this ensures high-quality data for our analysis, it does not fully capture the diversity and complexity of real-world clinical practice, which often involves fragmented, incomplete, or ambiguous information from electronic health records. Consequently, the observed performance—particularly on tasks that benefit from well-structured context—may reflect "textbook-style" reasoning and be inflated relative to real-world clinical environments. Third, model performance was primarily assessed using a metric defined as the degree of overlap between predicted multiple-choice answers and ground truth, which effectively corresponds to recall. This quantitative approach, while objective and allowing partial credit, does not capture several clinically important dimensions, including the relative severity of omissions (e.g., missing a critical diagnostic test vs. omitting a lower-yield option), the safety implications of incorrect or redundant selections, or the clinical appropriateness of the model's underlying reasoning. To better characterize omission-driven errors vs. over-selection of incorrect options, we additionally reported precision and F1 score in **Supplementary File 1**; however, more clinically grounded evaluations will require clinician-weighted or safety-oriented metrics that distinguish critical omissions from lower-impact errors. Fourth, because our evaluation is conducted in a closed-form multiple-choice setting in which models select from predefined options, it does not assess hallucinations or other failure modes that arise in open-ended content generation. Finally, our prompt engineering framework relied on static LLM-generated CoT examples and did not include real-time clinician feedback or reinforcement mechanisms, which may limit safety and adaptability in dynamic clinical environments.

**Implications for safe clinical integration of LLMs:** Together, this study demonstrates that strong LLM performance on medical knowledge exams does not necessarily translate to effective performance in sequential clinical decision-making tasks, as assessed within this set of structured clinical vignettes. This underscores a crucial gap between the theoretical knowledge of LLMs and their performance in more realistic clinical workflows. Our results further indicate that there is no "one-size-fits-all" solution for optimizing LLM performance in clinical decision-making; the effectiveness of prompting strategies to guide the language models is highly dependent on the specific clinical task at hand. This implies that the path toward safe and effective deployment of LLMs in healthcare involves the development of tailored approaches that pair the right model with a relevant prompting strategy to the right task, rather than a search for a single, universally optimal model or prompting technique. Accordingly, the safe clinical integration of LLMs will likely require hybrid, clinician-in-the-loop workflows with uncertainty-aware outputs, rather than fully autonomous use. In this setting, LLM recommendations merely serve as decision support and remain subject to human review and confirmation, particularly in ambiguous or high-risk cases. Future research can expand on this work by incorporating larger-scale and more diverse datasets that better reflect real-world clinical complexity, and evaluate whether newer reasoning models show similar task- and prompt-dependent behavior across sequential clinical decision-making workflows.

## Methods

**Data sources:** The dataset used in this study was sourced from the Merck Sharpe & Dohme (MSD) Clinical Manual Professional Version [11]. The MSD Manual is a widely recognized, evidence-based medical resource authored by subject matter experts and peer-reviewed prior to publication. It provides comprehensive clinical information across a wide range of specialties and is commonly used by healthcare professionals for education and decision support. A total of 36 clinical cases were available in this database at the time the study was conducted, each representing an individual patient case. The cases were extracted using web-scrapping in Python. These cases are structured to reflect realistic clinical scenarios and include key components of a standard patient encounter, such as HPI, ROS, PE results, and, where applicable, laboratory data.

**LLMs examined:** We evaluated the performance of three leading LLMs: two commercial models—ChatGPT-4o (OpenAI) and Gemini 2.5 Flash (Google)—and one open-source model—Llama 3.3 70B (Meta). Each model was accessed through its respective API. For ChatGPT-4o and Gemini 2.5 Flash, we used official API endpoints provided by OpenAI and Google, respectively. Requests were made using Python scripts through the `openai` and `google.generativeai` libraries. For Gemini 2.5 Flash, we disabled thinking by setting the thinking budget to zero (`thinkingBudget = 0`) to maintain comparability with the other evaluated models. For the open-source Llama 3.3 70B model, we employed the `together.ai` Python SDK, accessing the model via the `Together AI` platform. API requests were structured as `JSON` objects specifying the model's name, prompt messages, and function-calling parameters.

**Prompting LLMs:** Each clinical vignette was input into the model as a single continuous session to preserve contextual continuity. A new session was initialized for each case to avoid cross-contamination of responses between vignettes. Questions involving visual data, such as medical images, were excluded from analysis. Clinical decision-making tasks were ordered as follows: differential diagnosis, essential immediate steps, relevant diagnostic testing, final diagnosis, and treatment recommendation. These tasks were presented to the models as a series of prompts, where the input to the model for each clinical decision-making task included HPI, ROS, PE results, along with the ground truth for all preceding tasks. For example, model inputs for differential diagnosis were HPI, ROS, and PE results. The inputs to the next clinical task, essential immediate steps, included HPI, ROS, and PE results as well as the ground truth for differential diagnosis. Each question was tested three times (each in a new session) to account for potential variability in responses. The model's output for each trial was compared against a predefined set of correct answers extracted from the MSD database. Notably, not all case studies included all five question types; in such instances, the model was directed to proceed to the next available question type. The number of clinical vignettes for each category of questions is: 36 in essential differential diagnosis, 10 in essential immediate steps, 21 in relevant diagnostic testing, 33 in final diagnosis, and 31 in treatment ordering.

**Prompt engineering:** Prior to applying prompt engineering, we partitioned the cases for each task into training and test sets; approximately 30% of cases were allocated to the training set for the generation of few-shot examples with self-generated CoT reasoning, while the remaining cases formed the test set for model evaluation. We constructed the training dataset by identifying question instances for which the model's self-generated CoT reasoning produced an answer that matched the ground truth answer. Specifically, we prompted each LLM with the instruction "Let's think step by step." to elicit CoT reasoning. From the resulting output, we extracted the model's final answer and compared it to the correct answer for each question type. If the model's answer matched the ground truth, the case was included in the training set. Otherwise, it was assigned to the test set. In situations where the self-generated CoT did not yield a sufficient number of high-quality training examples, we supplied the correct answer to the model and prompted it to generate a corresponding CoT explanation. These synthetically augmented CoT examples were included in the training set until the desired sample size was achieved. Once the training set was finalized, we applied the structured MedPrompt prompting techniques to evaluate performance on the test set.

For each task, $K = 3$ cases from the training set were selected as examples using $K$NN or random dynamic few-shot prompting. For $K$NN dynamic few-shot learning introduced and utilized in MedPrompt [6], we first computed vector representations of both training and test set questions using the OpenAI's `client.embeddings.create()` function. Three few-shot examples were then selected for each test question based on cosine similarity, prioritizing examples that were semantically similar to the target test question in terms of clinical context. Answer to these questions already involved CoT explanation generated by the respective LLM under evaluation.

In addition to the original MedPrompt framework, which involved $K$NN dynamic few-short prompting [6], we also examined a variant of MedPrompt, where we replaced the $K$NN dynamic few-shot prompting with random dynamic few-shot prompting, where the examples from the training dataset for each test question were selected randomly for few-shot learning. For answer choice shuffling ensemble, another key component of MedPrompt, answer choices were shuffled to minimize answer-order bias. Three prompt variants with shuffled answer choices were generated and aggregated in an ensemble framework for robust evaluation.

**Performance evaluation:** Model performance for questions with multiple answer choices (differential diagnosis, essential immediate steps, relevant diagnostic testing, treatment recommendation) was calculated based on the percent overlap between the model-generated responses and the ground truth answers in the MSD database. This allowed us to award partial credit to LLMs when the model's answer intersected with the correct set but was not fully correct. For example, in differential diagnosis questions, if the ground truth listed five possible conditions (e.g., pneumonia, pulmonary embolism, asthma, heart failure, and bronchitis) and the model correctly identified three of them (e.g., pneumonia, asthma, and heart failure), then the performance score for that instance would be calculated as the proportion of correct predictions: 3 out of 5, or 60%. Each clinical question was evaluated over three independent runs to account for potential variability in model responses. In addition to the overlap-based accuracy score, which effectively corresponds to recall), we computed precision (proportion of selected options that were correct) and their harmonic mean (F1 score) for tasks with multiple correct answers. These metrics are defined as follows: Recall = $\frac{|P \cap G|}{|G|}$, Precision = $\frac{|P \cap G|}{|P|}$ , F1 = $\frac{2.|P \cap G|}{|P| + |G|}$, where $G$ and $P$ denote the set of correct answer choices (ground truth) and the set of answer choices selected by the model. These complementary metrics distinguish errors driven primarily by omissions (recall) vs. over-selection of incorrect options (precision) and are reported alongside the primary accuracy metric in **Supplementary File 1**.

For each case and run, accuracy (recall), precision, and F1-score were first calculated and the three run-level values were then averaged to obtain one case-level value for each metric. Overall performance for each question type was calculated as the mean of these case-level values across all relevant questions within that category.

For final diagnosis, where there was a single correct answer choice per case, the model's answer for each clinical case was directly compared with MSD correct answers, labeling 0 (incorrect) or 1 (correct). The average score was calculated for the three runs within each case, and overall accuracy was calculated across cases; for example, if the model answered correctly in two of three runs, the case-level accuracy was 2/3. Overall accuracy was then calculated as the mean of these case-level accuracies across all cases. Precision and F1 score were not reported because, in this single-answer setting, both are identical to accuracy at the case level and therefore also yield the same overall mean across cases.

For all tasks we have additionally reported 95% confidence intervals estimated by nonparametric bootstrapping over clinical vignettes (resampling vignettes with replacement and recomputing the mean performance). All baseline prompting, prompt engineering, and performance evaluation implementations were conducted within Google Colaboratory and can be accessed in **Supplementary File 2**.

**Statistical methods:** To evaluate the statistical significance of performance differences between LLMs under different prompting strategies or temperature settings, we conducted statistical hypothesis testing using the paired Wilcoxon Signed Rank test using case-level performance values. All tests were two-sided, and statistical significance was based on a p-value of < 0.05. All statistical analyses were performed using the `tidyverse`, `ggsignif` packages in R.

## Declarations

### Acknowledgements

The authors would like to acknowledge Kevin Wu and Yixuan Huang for their assistance in web-scraping clinical cases from the online database.

### Data availability

All data generated in this study along with the Google Colaboratory notebooks detailing all model evaluations for each LLM are provided in the supplementary materials.

### Author contributions

ARZ conceived the study. MC performed all model evaluations and analyses. CM and ARZ analyzed the results. CM and ARZ wrote the manuscript. All authors have read and approved the final manuscript.

### Conflict of interests

The authors declare no competing interests.

### Funding

This project was unfunded.

## Supplementary information

**Supplementary File 1**: Task-level LLM performance metrics under default and zero temperature settings and with MedPrompt-based prompting variants, along with p-values from pairwise statistical analyses.

**Supplementary File 2**: A compressed directory that includes the Google Colaboratory notebooks detailing the implementations of all analyses described in this study.